\documentclass[]{interact}

\usepackage{epstopdf}
\usepackage[caption=false]{subfig}
\usepackage{array}
\usepackage{float}
\usepackage[utf8]{inputenc}
\usepackage{multirow}
\usepackage{cite}
\usepackage{url}
\usepackage{xcolor}
\usepackage[normalem]{ulem}
\usepackage{soul}

\theoremstyle{plain}

\theoremstyle{definition}

\theoremstyle{remark}

\begin{document}

\title{Do Maps Still Matter for Machines: Revisiting the Role of Choropleth Maps in Foundation Model Spatial Understanding}


\author{
  Zhiwei Wei\textsuperscript{a,b}, 
  Yonghe Sun\textsuperscript{a,b}, 
  Zhenjia Liu\textsuperscript{a,b}, 
  Wenjia Xu\textsuperscript{c}, 
  Chao He\textsuperscript{d},
  Weihua Dong\textsuperscript{e}, 
  Chunbo Liu\textsuperscript{f}, 
  Hua Liao\textsuperscript{a,b*}\thanks{CONTACT Hua Liao. Email: liaohua@hunnu.edu.cn. ORCID: 0000-0002-6304-329}
  \\
  \affil{\textsuperscript{a}School of Geographic Sciences, Hunan Normal University, Changsha, China}
  \affil{\textsuperscript{b}Hunan Key Laboratory of Geospatial Big Data Mining and Application, Changsha, China}
  \affil{\textsuperscript{c}School of Information and Communication Engineering, Beijing University of Posts and Telecommunications, Beijing, China}
  \affil{\textsuperscript{d}School of Artificial Intelligence and Big Data, Wuhan Business University, Wuhan, China}
  \affil{\textsuperscript{e}Advanced Interdisciplinary Institute of Satellite Applications, State Key Laboratory of Earth Surface Processes and Resource Ecology, Faculty of Geographical Science, Beijing Normal University, Beijing, China}
  \affil{\textsuperscript{f}The Aerospace Information Research Institute, Chinese Academy of Sciences, Beijing, China}
}
\maketitle

\begin{abstract}
Spatial understanding is crucial for foundation models (FMs), and visual maps have long been a key tool for humans to understand space. Recent advances in FMs have dramatically improved their ability to process and reason over text, images, and other multimodal data, leading to a growing debate over whether traditional human-designed abstractions, such as maps, are still necessary for machine reasoning. This study revisits the role of choropleth maps in the spatial understanding of FMs, asking whether maps, as a long-standing cognitive medium for spatial understanding, still offer benefits in the era of multimodal models. To this end, we construct ChoroplethMap-Bench, a controlled benchmark consisting of 2,400 synthetic choropleth maps, corresponding GeoJSON data, and 12,000 questions, spanning five hierarchical cognitive dimensions: \textit{Identify}, \textit{Spatial Recognition}, \textit{Compare}, \textit{Rank}, and \textit{Delineate}. The benchmark compares three input conditions—\textbf{Data Only}, \textbf{Map Only}, and \textbf{Data + Map}—across 22 frontier models, including both open-source and proprietary systems. The results demonstrate that maps, particularly when combined with symbolic data, substantially enhance model performance, especially for tasks requiring higher-level spatial understanding. We further analyze how cartographic design factors, including map type, color hue, and spatial structure, influence reasoning performance. In addition, we examine language effects and response stability, showing the cross-lingual robustness and repeatability of modern multimodal models. Overall, our findings show that incorporating maps into FMs leads to improved geospatial reasoning accuracy, with the strongest gains observed under the \textbf{Data + Map} condition. This work highlights the continued relevance of maps as external representations for machine intelligence and suggests future directions for evaluating broader map forms and more complex reasoning tasks.
\end{abstract}


\section{Introduction}

Recent advances in foundation models (FMs) have substantially expanded the ability of machines to interpret the world directly \cite{wei2025thinking}. Contemporary models can process and reason over text, images, tables, diagrams, code, and increasingly complex multimodal environments \cite{Bommasani2021, Brown2020, OpenAI2023, Reed2022, awais2025foundation}. In many settings, machines are no longer limited to pre-designed symbolic pipelines, but can ingest raw information streams and generate decisions end-to-end. As machine intelligence moves closer to direct world understanding, a deeper question emerges: Does intelligence still benefit from human-designed abstractions, or can raw data alone suffice? This question is especially relevant for spatial information, where humans have long relied on maps to externalize geographic reasoning.

Among human-designed abstractions, maps are one of the most enduring and influential tools for organizing spatial knowledge \cite{WeiPanMap2025}. Rather than presenting the world as fragmented coordinates, tables, or textual descriptions, maps transform geographic complexity into structured visual forms that support reasoning about proximity, hierarchy, topology, regional disparity, and spatial pattern \cite{MacEachren1995, Montello2002, Slocum2009, Kraak2020}. Across domains such as urban planning, environmental management, epidemiology, transportation, and education, maps have long functioned not merely as illustrations, but as cognitive interfaces that reduce reasoning burden and reveal relationships difficult to infer from raw records alone \cite{Robinson1995, Fabrikant2009, Roth2013, Dong2019Gaze}. For humans, the value of maps is well established. Whether similar benefits extend to machines, however, remains largely unknown.

This uncertainty has become more important in the FM era. Modern models can directly consume structured geographic inputs such as regional statistics, tabular attributes, coordinate lists, textual reports, and machine-readable formats such as JSON \cite{yang2025mapcolorai, luo2026geojson}. In principle, such capabilities may reduce the need for traditional cartographic mediation. At the same time, growing literature has evaluated machine reasoning on charts, documents, natural images, embodied scenes, and navigation environments \cite{Masry2022, Mangalam2023, Majumdar2024,li2025does}. Yet most prior studies focus on perception or generic multimodal reasoning, rather than asking whether maps themselves remain useful representational interfaces for machines. \textbf{If a model can already process raw geographic data directly, does converting such data into maps still provide an advantage?} This seemingly straightforward question has, surprisingly, not been explored systematically.


To this end, we focus on choropleth maps—one of the most commonly used forms of thematic cartography, which convert raw numbers into interpretable spatial structures, enabling rapid perception of clusters, gradients, hotspots, boundaries, and anomalies \cite{Brewer1994, Brewer2002, Slocum2009, wei2026mapcoloragent}. They therefore can provide an ideal controlled testbed for comparing two fundamentally different information pathways for machine reasoning: direct access to raw data (e.g., GeoJSON) versus cartographic abstraction (maps). In this study, we construct a benchmark that evaluates machine performance across three input conditions—\textbf{Data Only}, \textbf{Map Only}, and \textbf{Data + Map}—using 2,400 synthetic choropleth maps, corresponding GeoJSON data, and 12,000 questions. The benchmark spans five cognitive dimensions: \textit{Identify}, \textit{Spatial Recognition}, \textit{Compare}, \textit{Rank}, and \textit{Delineate}, covering a range of tasks from local attribute reading to global spatial pattern recognition. Our results demonstrate that maps, especially when combined with data, enhance model performance, particularly in tasks that require more complex spatial reasoning. Even as FMs increasingly process raw data directly, maps continue to serve as valuable tools for improving machine intelligence, particularly in tasks that demand higher-level spatial understanding. Our main contributions are threefold:
\begin{itemize}
    \item We introduce \textbf{ChoroplethMap-Bench}, the first controlled benchmark for systematically evaluating whether maps improve FM spatial understanding. The benchmark contains 2,400 synthetic choropleth maps, corresponding GeoJSON data, and 12,000 questions spanning five hierarchical cognitive dimensions: \textit{Identify}, \textit{Spatial Recognition}, \textit{Compare}, \textit{Rank}, and \textit{Delineate}.

    \item Through systematic comparison under three representation conditions---\textbf{Data Only}, \textbf{Map Only}, and \textbf{Data + Map}---across 22 frontier foundation models, we show that maps still matter for machines: combining cartographic representations with symbolic data consistently outperforms raw data alone, with the largest gains observed in higher-level global reasoning tasks.

    \item We further characterize when maps help machine reasoning by analyzing map design factors (map type, color hue, and spatial structure), as well as robustness to prompting strategies, language variation, prior geographic knowledge, and repeated inference.
\end{itemize}

\section{Related work}

\subsection{Spatial Understanding in Foundation Models}

Recent years have witnessed rapid progress in foundation models (FMs), whose capabilities now extend beyond language generation to multimodal perception, reasoning, and interactive decision-making \cite{Bommasani2021, Brown2020, OpenAI2023}. As these models are increasingly deployed everywhere, evaluating their ability to understand spatial structure, reason about layouts, and operate under partial observability has become an important research direction.

Early studies mainly examined spatial reasoning in static visual scenes. Vision-language models were tested on relative position, object arrangement, perspective understanding, and geometric relationships in natural images \cite{liu2023visual, awais2025foundation}. In parallel, chart and diagram benchmarks demonstrated that models can partially reason over structured visual abstractions, although performance often declines when multi-step reasoning or precise quantitative comparison is required \cite{Masry2022, Methani2020, Kahou2018, li2025does}. These findings suggest that modern models can extract spatial cues from visual inputs, but their success depends strongly on representational clarity and task complexity.

A second research line focuses on dynamic or embodied spatial environments, in which models are asked to interpret egocentric videos, track motion, answer questions about changing scenes, or complete navigation-oriented tasks. Representative benchmarks such as OpenEQA \cite{Majumdar2024} reveal that long-horizon memory, self-localization, and action-grounded reasoning remain challenging. Related work in embodied AI and interactive agents further shows that spatial competence requires not only perception, but also planning, memory, and sequential decision-making \cite{Anderson2018, Reed2022, yang2025thinking, wei2025thinking}.

The rapid development of FMs has also attracted growing attention from the GIScience community, particularly under the emerging paradigm of GeoAI. As a result, a third research line has begun to explore geospatial intelligence, where FMs are applied to tasks such as remote sensing interpretation, geographic question answering, place understanding, spatial knowledge retrieval, and Earth observation analytics \cite{Mai2023, hong2024spectralgpt, ji2025foundation, zhang2025mapreader, liu2026can}. These studies highlight the growing potential of foundation models in geographic contexts. However, most existing work primarily emphasizes raw imagery, textual knowledge, or task-specific predictive pipelines. Comparatively little attention has been paid to whether cartographic representations themselves provide unique advantages as structured interfaces for machine spatial understanding.

Overall, existing literature establishes that FMs are increasingly capable of spatial reasoning across images, videos, embodied scenes, and geospatial data. However, prior work has primarily focused on what models can do, rather than how different representational forms influence what they can do. In particular, the role of maps as structured interfaces for machine reasoning remains insufficiently examined.

\subsection{Maps as External Spatial Representations for Intelligence}

Long before the emergence of machine intelligence, maps had already become one of humanity’s most successful external representations for organizing spatial knowledge \cite{WeiPanMap2025}. By selectively abstracting the real world into symbols, geometry, and visual structure, maps enable efficient reasoning about distance, adjacency, region, direction, and spatial pattern \cite{Robinson1995, MacEachren1995, Slocum2009}. In this sense, maps are not merely descriptive graphics, but cognitive tools that transform geographic complexity into interpretable forms \cite{Dong2019Gaze}.

Research in cartography and spatial cognition has long documented how maps support human reasoning. Cognitive studies show that map design influences spatial memory, search efficiency, route planning, and pattern recognition \cite{Montello2002, Fabrikant2009, dong2020does, yang2026learn}. Geovisualization research further demonstrates that appropriately designed maps can reveal hidden structures in complex data, especially when users must compare regions, identify clusters, or detect trends across space \cite{Andrienko2006, Roth2013}. These findings suggest that maps function as external cognitive scaffolds rather than passive displays.

Among map types, thematic maps are especially relevant for information reasoning because they integrate spatial structure with statistical attributes. Choropleth maps, in particular, represent regional values through area-based color encoding and are widely used in demography, epidemiology, economics, climate science, and election communication \cite{Brewer1994, Kraak2020, yang2025mapcolorai}. Extensive cartographic research has investigated classification methods, color schemes, perceptual bias, normalization effects, and uncertainty communication in choropleth design \cite{Jenks1967, Brewer2002, Olson1976, Krygier2016, wu2024computational}. This literature largely focuses on improving human interpretability, showing that visual design choices strongly affect the ability to infer clusters, gradients, and outliers.

With the rise of artificial intelligence, maps have also begun to appear as machine inputs in applications such as map parsing, road extraction, map-image retrieval, and multimodal geographic systems \cite{Chiang2014, martinez2023deep, petitpierre2024fragment, zhou2024cartomark, zhang2025mapreader}. However, these studies typically treat maps as another data modality for downstream tasks, rather than asking a more fundamental question: do maps remain useful abstractions when machines can already process raw structured data directly? This distinction is important. If modern FMs can reason equally well from tables, coordinates, or JSON records, maps may be redundant for machines. Conversely, if map-based abstraction still improves performance, then maps continue to serve as efficient external interfaces not only for humans, but also for machine intelligence. To our knowledge, this representational question has not been systematically investigated under controlled benchmark settings. Our work addresses this gap through a comparative study based on choropleth maps.

\section{Methodology}
\subsection{Overview of the Framework}

\begin{figure} [H]
	\centering
	\includegraphics[width=\textwidth]{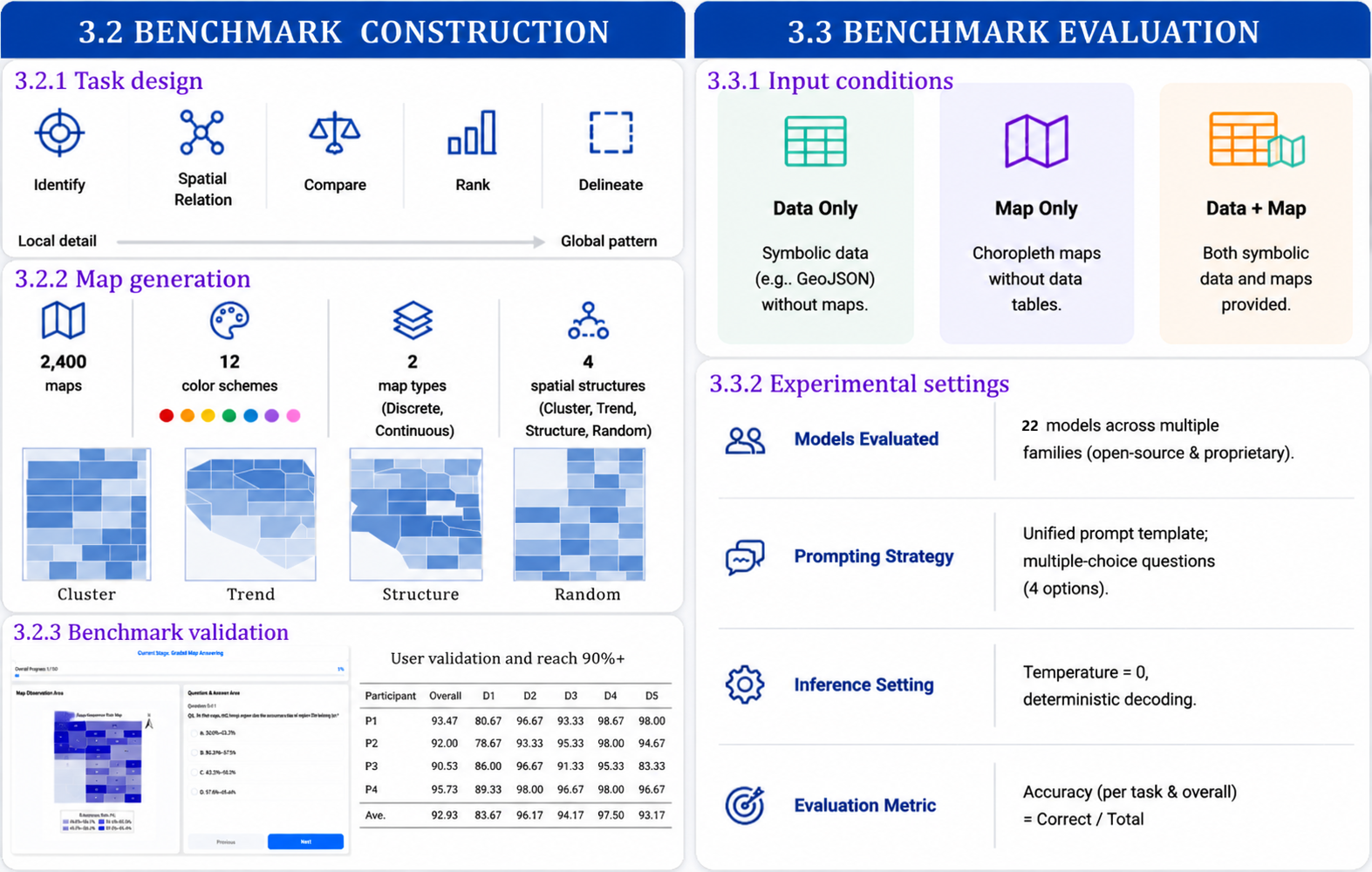}
	\caption{Overview of the ChoroplethMap-Bench construction and evaluation framework. The benchmark contains 2,400 synthetic choropleth maps, their corresponding GeoJSON data, and 12,000 multiple-choice questions covering five spatial-understanding dimensions. Foundation models are evaluated under three input conditions—Data + Map, Data Only, and Map Only—and the results are further analyzed with respect to model performance, cartographic design factors, prompting strategies, language, geographic context, decoding settings, classification methods, and response stability.}
	\label{framework}
\end{figure}

The overall framework is illustrated in Figure \ref{framework}. Our goal is to systematically examine whether choropleth maps still provide meaningful benefits for FMs when compared with direct access to raw geodata. To this end, the framework is organized into two connected components: \textbf{benchmark construction} and \textbf{benchmark evaluation}.

First, we build ChoroplethMap-Bench, a controlled dataset designed for fair comparison between symbolic geographic data and cartographic representations. Specifically, each benchmark instance contains a one-to-one correspondence between structured GeoJSON data (symbolic geographic data) and its rendered choropleth map (cartographic representation). We generate 2,400 synthetic maps under controlled variations of map type, color hue, and spatial structure. Based on these maps, 12,000 multiple-choice questions are automatically produced, covering five hierarchical cognitive dimensions: \textit{Identify}, \textit{Spatial Recognition}, \textit{Compare}, \textit{Rank}, and \textit{Delineate}, enabling evaluation from local attribute retrieval to higher-level global pattern reasoning.

Second, the same benchmark instances are evaluated under three representation conditions across 22 frontier FMs: \textbf{Data Only}, in which models receive only structured GeoJSON inputs; \textbf{Map Only}, in which models receive only choropleth map images; and \textbf{Data + Map}, in which symbolic data and maps are jointly provided. Since all conditions share identical underlying tasks and ground-truth answers, performance differences can be directly attributed to representational format.

Based on the evaluation framework, we further analyze how maps influence machine reasoning across different settings, including overall task performance, map design factors, prompting strategies, language variation, prior geographic knowledge, and repeated inference stability.

\subsection{ChoroplethMap-Benchmark Construction}
\subsubsection{Tasks design}

The tasks in ChoroplethMap-Bench are deliberately designed under a hierarchical framework that progresses from local perception to global pattern understanding, which mirrors how humans gradually build spatial understanding.\cite{Montello2002, Golledge1999, Roth2013}. Following this principle, the benchmark organizes questions into five task dimensions: \textit{Identify}, \textit{Spatial Recognition}, \textit{Compare}, \textit{Rank}, and \textit{Delineate}. Lower-level dimensions, such as \textit{Identify} and \textit{Spatial Recognition}, mainly evaluate local retrieval and basic relational understanding. Intermediate dimensions, such as \textit{Compare}, focus on attribute-based reasoning between regions. Higher-level dimensions, including \textit{Rank} and \textit{Delineate}, require broader reasoning over global spatial distributions and structural patterns, where visual compression is expected to provide the greatest advantage.

For each benchmark map, we generate one multiple-choice question for each task dimension. Therefore, every map is paired with five questions in total. Within these dimensions, multiple question subtypes are further designed, resulting in 12 fine-grained question templates overall (Q1-Q12). All questions are multiple-choice questions, but the number of answer options varies across task dimensions and question subtypes. Specifically, D3 and some D4 questions contain three options, while some D5 questions contain two options; the remaining questions contain four options. Ground-truth answers are generated deterministically from the map geometry and attribute values. Detailed task dimensions, question subtypes, and generated question forms are summarized in Table \ref{tab:task-details}.

\begin{table}[H]
\centering
\small
\caption{Five spatial-understanding dimensions and 12 question subtypes included in ChoroplethMap-Bench.}
\label{tab:task-details}
\renewcommand{\arraystretch}{1.25}
\setlength{\tabcolsep}{5pt}

\begin{tabular}{>{\centering\arraybackslash}m{0.12\textwidth}
                >{\centering\arraybackslash}m{0.24\textwidth}
                >{\arraybackslash}m{0.56\textwidth}}
\toprule
\textbf{Task dimension} & \textbf{Question Subtype} & \textbf{Question Description} \\
\midrule

\multirow{2}{=}{\centering D1. Identify}
& attr2region
& Q1. Given an attribute level, identify the corresponding region. \\
& region2attr
& Q2. Given a specific region, identify which attribute level it belongs to. \\
\midrule

\multirow{2}{=}{\centering D2. Spatial Recognition}
& Direction recognition
& Q3. Determine the directional spatial relationship between two regions. \\
& Adjacent recognition
& Q4. Determine whether two regions are adjacent or non-adjacent. \\
\midrule

\centering D3. Compare
& Attribute comparison
& Q5. Compare the attribute values of two given regions and determine which is higher, lower, or equal. \\
\midrule

\multirow{2}{=}{\centering D4. Rank}
& Global rank
& Q6. Find the region with the highest (or lowest) attribute value across the entire map. \\
& Local rank
& Q7. Find the region with the highest (or lowest) attribute value among the neighbors of a given region. \\
\midrule

\multirow{5}{=}{\centering D5. Delineate}
& \multirow{2}{=}{\centering Cluster delineate}
& Q8. Determine the number of spatial clusters. \\
&
& Q9. Determine whether a given region belongs to a cluster. \\

& Trend delineate
& Q10. Identify the direction of a monotonic gradient (trend) in the attribute distribution. \\

& \multirow{2}{=}{\centering Structure delineate}
& Q11. Determine the number of ring structures. \\
&
& Q12. Identify regions located on or inside a ring. \\

\bottomrule
\end{tabular}
\end{table}

\subsubsection{Map generation}

To support the benchmark tasks described above, we generated a controlled set of synthetic choropleth maps. The generation design follows two considerations. First, because choropleth maps may vary in visual encoding, we explicitly control two cartographic design factors: \textit{map type} and \textit{color hue}. Second, because several benchmark tasks (Q8-Q12 in Table \ref{tab:task-details}) require recognizing spatial clusters, trends, and structures, we also control the \textit{spatial structure} of attribute distributions. This design allows ChoroplethMap-Bench to evaluate not only whether maps help FMs but also how different map designs and spatial patterns influence model performance. The details of the map generation are illustrated as follows.

\textbf{(1) Map type.}
We consider two widely used choropleth forms: discrete and continuous, which represent two fundamentally different strategies for encoding geographic attributes \cite{Brewer1994}. The terms discrete and continuous refer exclusively to color encoding schemes rather than the distinction between discrete and continuous geographic phenomena in GIScience. In both cases, the underlying geographic entities are identical polygonal administrative regions. For discrete maps, continuous attribute values are grouped into ordered classes and visualized using distinct color levels. In this study, we adopt four classes, as this is a common practical compromise in choropleth design: it preserves sufficient ordinal differentiation while maintaining clear perceptual separability and avoiding excessive legend complexity \cite{Brewer1994, yang2025mapcolorai}. For continuous maps, raw attribute values are preserved and encoded through smooth sequential color variation without explicit class boundaries.

\textbf{(2) Color hue.}
In practical choropleth mapping, too many hues are usually avoided because quantitative values are intended to reflect ordered magnitude rather than categorical differences \cite{Brewer1994, yang2025mapcolorai}. Therefore, each map in our benchmark adopts a single base hue, while attribute variation is primarily expressed through changes in brightness or saturation. To prevent the benchmark from being tied to any specific color preference of foundation models, we further cycle through 12 evenly spaced hue settings in HSV space across all benchmark maps. For discrete maps, the 12 hues are cycled in HSV space. For each hue, two visual encoding modes are used equally often. In the saturation-variable mode, saturation is set to $S=[0.20, 0.40, 0.60, 0.95]$, while brightness is randomly sampled from $B \in [0.65, 0.85]$. In the brightness-variable mode, brightness is set to $B=[0.75, 0.60, 0.45, 0.30]$, while saturation is randomly sampled from $S \in [0.45, 0.60]$. For continuous maps, each attribute value is first normalized as $x=(V-V_{\min})/(V_{\max}-V_{\min})$. Then either saturation or brightness varies linearly with $x$: in the saturation-variable mode, $S=0.10+0.85x$ with brightness fixed randomly in $[0.40,0.60]$; in the brightness-variable mode, $B=0.75-0.45x$ with saturation fixed randomly in $[0.40,0.60]$.

\begin{figure}[H]
	\centering
	\includegraphics[width=\textwidth]{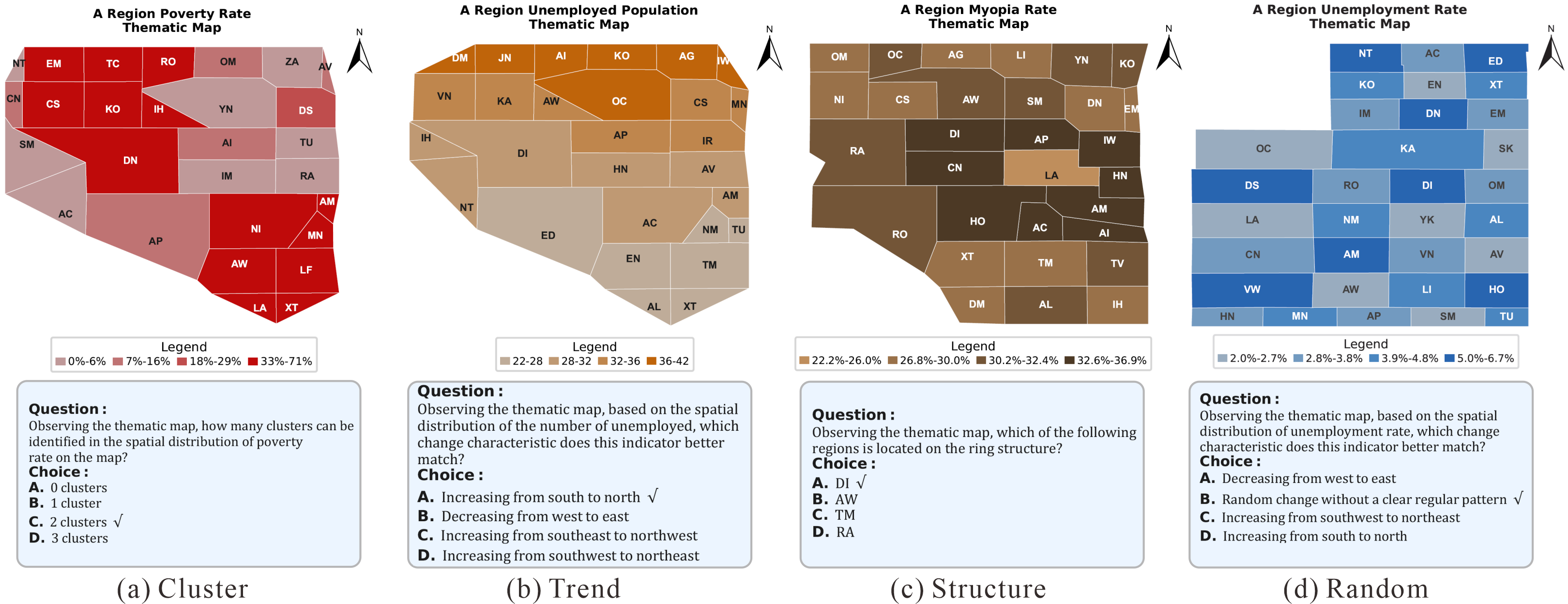}
	\caption{Representative choropleth maps generated under four spatial-distribution patterns: (a) Cluster, in which adjacent regions form spatially concentrated high- or low-value groups; (b) Trend, characterized by a directional attribute gradient; (c) Structure, containing ring-shaped spatial configurations; and (d) Random, with no predefined global spatial organization. Each panel also presents an example benchmark question associated with the corresponding spatial pattern.}
	\label{Design}
\end{figure}

\textbf{(3) Spatial structure.}
We control four spatial structures because benchmark questions in Table \ref{tab:task-details} explicitly require recognition of clustered, directional, and structural spatial organizations. Accordingly, we include three organized patterns---\textit{cluster}, \textit{trend}, and \textit{structure}---together with an additional \textit{random} pattern that serves as an unstructured baseline without salient global organization. For each valid clipped region set, one structure is assigned according to a balanced cycle; the examples are depicted in Figure \ref{Design}. 

\begin{itemize}
    \item \textbf{Cluster:} For \textit{cluster} maps, an adjacency dictionary is used to expand from randomly selected seed regions and form up to two compact high-value or low-value clusters. Core regions are assigned extreme class values, while surrounding buffer regions receive intermediate values.

    \item \textbf{Trend:} For \textit{trend} maps, a direction is randomly selected from $0^\circ$, $45^\circ$, $90^\circ$, or $135^\circ$. Region centroids are projected onto this direction, and discrete class values are assigned by quantile cuts along the projection. Continuous values are assigned monotonically following the selected trend direction.

    \item \textbf{Structure:} For \textit{structure} maps, one or two core regions are selected near the map center, and circular buffers are generated to create ring-like spatial structures. Ring regions receive one extreme class value, whereas core regions receive the opposite extreme value.

    \item \textbf{Random:} For \textit{random} maps, attribute values are assigned randomly to regions without enforcing any predefined spatial regularity. As a result, the map does not exhibit clear clusters, directional gradients, or organized structural patterns, and serves as an unstructured baseline for evaluating model reasoning under irregular spatial distributions.
\end{itemize}

\textbf{(4) Generation process.}
All maps are generated through a fully automated and reproducible Python pipeline using \texttt{geopandas} and \texttt{matplotlib}. The generation process consists of five main steps: geographic sampling, spatial preprocessing, attribute assignment, symbolic data construction, and map rendering.

\begin{itemize}
    \item \textbf{Step 1: Geographic sampling.}  Base geographic data are loaded from county-level administrative shapefiles of multiple countries, such as the United States, France, Germany, and Switzerland. We intentionally adopt county-level vector geometries because these units are smaller and more numerous than higher-level administrative regions, making it more difficult for models to infer place identities from familiar outlines. For each benchmark instance, a country shapefile is randomly selected and reprojected. A square clipping window with side length proportional to the country bounding box is randomly placed, and only windows containing 15--40 regions are retained.
    \item \textbf{Step 2: Spatial preprocessing.}  To ensure visual clarity and satisfy model input constraints, all MultiPolygons are exploded into individual Polygons. Containment pairs are detected and resolved by subtraction to avoid visual nesting. TopoJSON shared-arc simplification with adaptive epsilon is then applied to reduce the maximum number of exterior vertices per polygon to 8. If this constraint is not satisfied after initial simplification, an exponential search followed by binary refinement is used to identify the smallest epsilon that satisfies the vertex limit. We adopt exponential search because the minimum feasible simplification threshold is unknown a priori. Exponential search rapidly identifies an upper bound that satisfies the vertex constraint with only a logarithmic number of evaluations. Once such an upper bound is found, binary refinement is performed within the identified interval to efficiently determine the smallest epsilon that satisfies the vertex constraint while preserving as much geometric detail as possible. Additional quality filters are also applied: regions with areas below a country-specific threshold are removed, and polygons with compactness $(4\pi A/P^2)$, where $A$ and $P$ denote the polygon area and perimeter, respectively, lower than 0.1 are discarded to eliminate narrow edge fragments. Compactness is adopted because it is a widely used geometric descriptor for identifying highly elongated or fragmented polygons. Removing polygons with extremely low compactness helps eliminate thin edge fragments introduced by clipping and simplification while preserving the overall spatial structure of the benchmark.
    \item \textbf{Step 3: Attribute assignment.}  After the spatial subset is determined, attribute values are assigned according to the selected map pattern. Rather than generating arbitrary synthetic numbers, the benchmark draws value distributions from real U.S. statistical variables provided in the MapQA dataset \cite{chang2022mapqa}. These values are then randomly allocated to the selected regions while preserving the intended spatial pattern (e.g., cluster, trend, structure, or random). This strategy ensures that benchmark attributes remain realistic in magnitude and distribution while minimizing semantic priors tied to specific real-world regions.
    \item \textbf{Step 4: Symbolic data construction.}  A compact GeoJSON file is generated for each map, recording region geometries, attribute values, and spatial metadata. To further reduce semantic priors, each region is assigned a randomly generated two-letter English code (e.g., AB, QF, MT) as its name. A \texttt{geo\_origin} offset is applied to shorten coordinate digits, rings are downsampled, and coordinates are rounded to four decimal places. The file is further compressed through a multi-strategy pipeline to keep the symbolic input compact.
    \item \textbf{Step 5: Map rendering.} Each map is rendered at $640 \times 640$ pixels with 300 DPI. Region labels are placed at centroids with area-scaled font size. Label color is automatically selected using a luminance-based contrast rule: white text is used when the background luminance is $\leq 0.52$, and dark gray (\#1A1A1A) is used otherwise. A north arrow is added in the upper-right corner, and the title and legend unit are automatically inferred from the JSON class descriptions.
\end{itemize}

In total, ChoroplethMap-Bench contains 2,400 synthetic choropleth maps, including 1,200 discrete 4-class maps and 1,200 continuous maps. The maps are evenly distributed across four spatial structures—\textit{cluster}, \textit{trend}, \textit{structure}, and \textit{random}—with 600 maps for each structure. The benchmark also cycles through 12 hue settings in HSV space, so that the evaluation is not tied to a single color palette. Every benchmark sample forms a strict one-to-one correspondence between symbolic GeoJSON data and rendered map imagery, ensuring fair comparison across all input conditions. Representative examples of the four spatial structures are shown in Figure \ref{Design}.





\subsubsection{Benchmark validation}








To examine whether the generated benchmark is visually interpretable and logically valid, we conducted a human validation study on a randomly sampled subset of ChoroplethMap-Bench. Specifically, 25\% of the benchmark maps (600 maps in total) were randomly selected, and participants answered the same multiple-choice questions used in the benchmark setting. The validation involved four third-year undergraduate students majoring in Geographic Information Science. All participants were male, approximately 20 years old, and had completed formal coursework in Cartography, Geographic Information Systems, and Spatial Analysis. The results are shown in Table \ref{tab:human_validation}. We can observe that human participants achieved a high average overall accuracy of 92.93\%, indicating that the generated maps and associated questions are generally clear and solvable. Across the five task dimensions, the highest average accuracy was observed on \textit{D4 Rank} (97.50\%) and \textit{D2 Spatial Recognition} (96.17\%), suggesting that participants could reliably interpret local relations and identify extrema from choropleth maps. Comparatively lower accuracy was found on \textit{D1 Identify} (83.67\%) and \textit{D5 Delineate} (93.17\%), which mainly involve direct value decoding and higher-order global pattern recognition. Overall, these results confirm that ChoroplethMap-Bench is visually reasonable, logically consistent, and suitable for evaluating machine spatial understanding.

\begin{table}[H]
\centering
\small
\captionsetup{justification=centering}  
\caption{Human validation results on a randomly sampled subset of ChoroplethMap-Bench (\%).}
\label{tab:human_validation}
\renewcommand{\arraystretch}{1.12}
\setlength{\tabcolsep}{4.5pt}

\begin{tabular}{lcccccc}
\toprule
Participant & Overall & D1 & D2 & D3 & D4 & D5 \\
\midrule
P1   & 93.47 & 80.67 & 96.67 & 93.33 & 98.67 & 98.00 \\
P2   & 92.00 & 78.67 & 93.33 & 95.33 & 98.00 & 94.67 \\
P3   & 90.53 & 86.00 & 96.67 & 91.33 & 95.33 & 83.33 \\
P4   & 95.73 & 89.33 & 98.00 & 96.67 & 98.00 & 96.67 \\
\midrule
Ave. & 92.93 & 83.67 & 96.17 & 94.17 & 97.50 & 93.17 \\
\bottomrule
\end{tabular}
\end{table}

\subsection{ChoroplethMap-Bench Evaluation}


\subsubsection{Input conditions}

For each benchmark instance, we construct three input conditions to directly evaluate the effectiveness of maps as spatial representations for FMs. As the symbolic counterpart, we use GeoJSON because it is a widely adopted machine-readable geospatial format that explicitly preserves region geometries, coordinates, and attribute values without cartographic abstraction. Specifically, we compare GeoJSON input, map input, and their combined form under the same benchmark tasks:

\begin{itemize}
    \item \textbf{Symbolic-only input (Data Only)}: Models receive only the structured GeoJSON representation. This condition serves as the baseline for evaluating pure symbolic geospatial reasoning. Discrete maps include class indices and textual class descriptions, while continuous maps contain raw numerical values without discrete classes.
    \item \textbf{Combined inputs (Data + Map)}: In addition to the GeoJSON, models are provided with the corresponding choropleth map image. The map faithfully visualizes the same spatial semantics encoded in the JSON, serving as visual grounding. This is the core experimental condition used to quantify the cognitive compression effect induced by maps.
    \item \textbf{Map-only input (Map Only)}: Models receive only the generated choropleth map image, without access to any symbolic coordinate or attribute data. This condition evaluates whether models can infer spatial relations directly from visual representations alone.
\end{itemize}

By comparing model performance across these three conditions while holding the underlying tasks constant, performance differences can be attributed to representation format rather than task content. This controlled design enables a direct assessment of whether and when maps improve FMs' geospatial reasoning.

\subsubsection{Experimental settings}

\hspace*{1em}\textbf{(1) Evaluated models.}  
To systematically examine the cognitive compression effect of choropleth maps on FMs' geospatial reasoning, we evaluate a diverse set of recent multimodal large language models from both locally deployable and API-accessed ecosystems, spanning different model scales, architectures, and vision-language pretraining paradigms. All API-based experiments were conducted between \textbf{February 18 and April 28, 2026}. The selected models represent some of the strongest and most competitive publicly accessible systems available during this evaluation period.
Specifically, the locally deployable models include representative families such as Qwen3.5 (2B/4B/9B), Qwen3-VL (2B/4B/8B-Instruct), Qwen2.5-VL (3B/7B-Instruct), GLM-4.6V-Flash, InternVL3.5 (2B/4B/8B), and Gemma-3 (4B/12B), covering a wide range of architectures and parameter scales from a few billion to approximately 10 billion parameters. The API-accessed models include MiMo-V2-Omni, Gemini-2.5-Flash, Qwen3.6-Plus, Doubao-Seed-2.0-lite, Kimi-K2.5, GPT-5.5, Gemini-3.5-Flash, and Claude Opus 4.8. These models exhibit substantially more heterogeneous and generally larger effective capacities, although exact parameter counts are not fully disclosed for most proprietary systems. Among them, publicly reported estimates suggest that these models may operate at significantly larger scales (up to the trillion-parameter level in mixture-of-experts configurations). In particular, Kimi-K2.5 is an open-weight model; however, we evaluate it under the same API-based inference setting as proprietary models to ensure a consistent and reproducible experimental pipeline.

\textbf{(2) Evaluation protocol.}  
For each benchmark instance under each input condition, models are prompted in a zero-shot setting to generate a single answer for the target multiple-choice question. To ensure fair comparison, the same prompting template is used across all models and all input conditions. Model outputs are automatically parsed and matched against the benchmark ground-truth labels. Each model is evaluated independently under identical benchmark settings.

All experiments involving the locally deployable models are conducted on a dual-GPU server equipped with two NVIDIA RTX 4090 (24GB) GPUs and 64GB of system RAM. To process the full benchmark efficiently, we employ two-GPU parallel inference with 4-bit quantization for memory efficiency. For larger models, part of the model weights are automatically offloaded to system RAM by the inference framework when GPU memory is insufficient. The maximum generation length is set to 64 new tokens. Initial decoding uses greedy inference for deterministic outputs. If the returned format is invalid, a retry mechanism is triggered using temperature $=0.65$ and top-$p=0.92$, with at most two retries. Memory usage is actively managed through cache release and garbage collection during large-scale evaluation. Proprietary models are evaluated through stable public APIs that support joint image--text input. Specifically, GPT-5.5, Claude Opus 4.8, and Gemini models are accessed via OpenRouter, while China-based proprietary models (e.g., Doubao-Seed-2.0-lite, Qwen3.6-Plus, MiMo-V2-Omni, and Kimi-K2.5) are accessed through their official APIs.

\textbf{(3) Evaluation metric.}  
Model performance is measured using accuracy, defined as the proportion of correctly answered instances among all evaluated questions.

\section{Result}

To assess whether the three input settings (Data+Map, Data Only, and Map Only) exhibit statistically significant performance differences across the evaluated models, we first examined the normality of the overall accuracies using the Shapiro--Wilk test. Since at least one group violated the normality assumption ($p<0.05$), we adopted the non-parametric Friedman test for repeated measures. Post hoc pairwise comparisons were subsequently conducted using the Nemenyi test.

Statistical analysis confirms that the three input settings produce significantly different performance across the evaluated models. The Friedman test reveals a significant overall effect ($\chi^2=33.09, p<0.001$), with a large effect size ($f=0.44$). Post hoc Nemenyi comparisons further show that Data+Map significantly outperforms both Data Only ($p=0.003$) and Map Only ($p=0.001$), while Data Only also significantly outperforms Map Only ($p=0.042$). These results statistically support the conclusion that combining symbolic data with rendered maps consistently yields the strongest spatial reasoning performance.

Because the number of answer options varies across question subtypes, the chance probability is not uniformly 25\%. Specifically, D3 and some D4 questions have three options, some D5 questions have two options, and the remaining questions have four options. Accordingly, the chance probability for each question was defined as the reciprocal of its number of available options, resulting in an option-count-weighted aggregate chance level of 27.57\%. To determine whether model performance exceeded random guessing, we conducted one-sided Poisson-binomial tests over all 12,000 benchmark questions under each input condition, treating the questions as independent Bernoulli trials with question-specific success probabilities of $1/2$, $1/3$, or $1/4$. The results show that 65 of the 66 model--input condition combinations performed significantly above chance (all $p<0.001$). The only exception was Qwen3-VL-2B-Instruct under the Data Only condition, for which the observed accuracy was 28.11\% and did not significantly exceed chance ($p=0.092$). Thus, with this single exception, the evaluated models demonstrated performance reliably above the chance level across the tested input conditions.






\subsection{Overall Performance}

We evaluate all models under three distinct input conditions: \textbf{Data + Map}, \textbf{Data Only}, and \textbf{Map Only}, and the experimental results are summarized in Table \ref{tab:overall-results}. The results clearly indicate that the \textbf{Data + Map} condition consistently yields the highest accuracy for nearly all models evaluated (20 out of 22), demonstrating that the combination of symbolic data and visual maps is crucial for enhancing spatial reasoning capabilities. The key observations are summarized as follows.

\begin{table*}[htbp]
\centering
\small
\setlength{\tabcolsep}{4pt}   
\caption{Accuracy (\%) of 22 foundation models on ChoroplethMap-Bench under the Data + Map, Data Only, and Map Only input conditions. Overall denotes accuracy across all benchmark questions, while D1–D5 denote Identify, Spatial Recognition, Compare, Rank, and Delineate, respectively. The best model result in each column is shown in bold. Human-validation results are included as a reference. }
\label{tab:overall-results}

\makebox[\textwidth][c]{
\begin{tabular}{l 
                >{\centering\arraybackslash}p{0.85cm}
                >{\centering\arraybackslash}p{0.47cm}
                >{\centering\arraybackslash}p{0.47cm}
                >{\centering\arraybackslash}p{0.47cm}
                >{\centering\arraybackslash}p{0.47cm}
                >{\centering\arraybackslash}p{0.47cm}
                >{\centering\arraybackslash}p{0.85cm}
                >{\centering\arraybackslash}p{0.47cm}
                >{\centering\arraybackslash}p{0.47cm}
                >{\centering\arraybackslash}p{0.47cm}
                >{\centering\arraybackslash}p{0.47cm}
                >{\centering\arraybackslash}p{0.47cm}
                >{\centering\arraybackslash}p{0.85cm}
                >{\centering\arraybackslash}p{0.47cm}
                >{\centering\arraybackslash}p{0.47cm}
                >{\centering\arraybackslash}p{0.47cm}
                >{\centering\arraybackslash}p{0.47cm}
                >{\centering\arraybackslash}p{0.47cm}} 
\toprule

\multirow{2}{*}{\textbf{Model}} 
& \multicolumn{6}{c}{\textbf{Accuracy (Data + Map)}}  
& \multicolumn{6}{c}{\textbf{Accuracy (Data Only)}} 
& \multicolumn{6}{c}{\textbf{Accuracy (Map Only)}} \\
\cmidrule(lr){2-7} \cmidrule(lr){8-13} \cmidrule(lr){14-19}
& Overall & D1 & D2 & D3 & D4 & D5   
& Overall & D1 & D2 & D3 & D4 & D5 
& Overall & D1 & D2 & D3 & D4 & D5  \\
\midrule

Qwen3.5-9B          & 61.3 & 80.8 & 58.4 & 62.2 & 70.2 & 35.0 & 52.6 & 70.0 & 43.3 & 58.8 & 60.7 & 30.2 & 42.7 & 43.5 & 51.8 & 43.2 & 40.6 & 34.5 \\
Qwen3.5-4B          & 60.8 & 70.8 & 62.5 & 59.7 & 71.1 & 40.0 & 47.2 & 55.0 & 40.6 & 56.6 & 57.1 & 27.0 & 47.4 & 43.5 & 66.0 & 44.9 & 43.7 & 39.3 \\
Qwen3.5-2B          & 43.4 & 38.2 & 52.2 & 49.8 & 56.7 & 20.4 & 37.4 & 27.5 & 33.4 & 47.4 & 56.4 & 22.5 & 37.2 & 33.3 & 58.0 & 37.5 & 36.5 & 20.7  \\
Qwen3-VL-8B-Instruct & 40.6 & 50.0 & 37.1 & 31.7 & 52.4 & 32.2 & 37.4 & 42.9 & 31.6 & 30.8 & 52.6 & 29.4 & 34.8 & 34.5 & 44.6 & 28.1 & 36.4 & 30.8  \\
Qwen3-VL-4B-Instruct & 33.7 & 50.0 & 27.3 & 26.6 & 45.8 & 19.1 & 32.3 & 44.7 & 24.4 & 29.7 & 44.7 & 18.3 & 30.6 & 33.7 & 38.4 & 30.9 & 31.0 & 18.9  \\
Qwen3-VL-2B-Instruct & 29.6 & 31.2 & 24.8 & 31.7 & 34.8 & 25.6 & 28.1 & 31.1 & 24.2 & 31.6 & 29.9 & 23.9 & 29.2 & 28.8 & 27.3 & 32.1 & 29.4 & 28.6  \\
Qwen2.5-VL-7B-Instruct & 39.9 & 46.4 & 35.1 & 34.4 & 56.3 & 27.4 & 33.3 & 33.1 & 29.3 & 34.4 & 43.9 & 25.9 & 34.1 & 31.8 & 38.3 & 33.2 & 36.1 & 31.2  \\
Qwen2.5-VL-3B-Instruct & 31.9 & 39.6 & 28.3 & 30.5 & 37.0 & 24.4 & 29.8 & 32.8 & 26.5 & 29.7 & 37.1 & 23.2 & 29.5 & 32.9 & 32.3 & 29.6 & 29.2 & 23.6  \\
GLM-4.6V-Flash & 64.2 & 73.2 & 62.1 & 64.7 & 85.8 & 35.5 & 40.3 & 61.1 & 29.3 & 36.2 & 46.6 & 28.4 & 50.7 & 46.4 & 67.5 & 42.9 & 59.0 & 37.7  \\
InternVL3.5-8B & 64.9 & 73.2 & 54.2 & 65.7 & 91.0 & 40.4 & 60.8 & 67.3 & 46.1 & 65.8 & 92.0 & 32.9 & 46.9 & 41.6 & 63.1 & 40.7 & 53.1 & 32.9 \\
InternVL3.5-4B & 61.0 & 66.7 & 58.3 & 65.6 & 80.9 & 33.8 & 52.3 & 53.0 & 43.2 & 61.1 & 80.1 & 24.0 & 47.7 & 43.0 & 61.2 & 45.6 & 51.8 & 37.0 \\
InternVL3.5-2B & 43.1 & 40.9 & 39.3 & 44.4 & 58.6 & 32.5 & 39.6 & 32.4 & 32.2 & 43.8 & 56.0 & 33.5 & 35.9 & 31.4 & 42.5 & 36.2 & 37.9 & 31.9 \\
Gemma3-12B-IT & 50.6 & 56.9 & 52.3 & 52.4 & 62.4 & 29.2 & 48.0 & 55.3 & 35.8 & 53.7 & 66.7 & 28.5 & 30.3 & 26.7 & 37.8 & 32.4 & 26.8 & 27.8  \\
Gemma3-4B-IT & 33.4 & 27.5 & 31.6 & 33.8 & 39.6 & 34.5 & 32.6 & 26.4 & 30.1 & 33.8 & 40.2 & 32.3 & 28.9 & 27.8 & 24.7 & 32.8 & 26.9 & 32.2 \\
\midrule
MiMo-V2-Omni & 67.4 & 92.7 & 56.4 & 67.3 & 82.9 & 38.0 & 66.6 & 92.6 & 53.6 & 65.2 & 86.8 & 34.6 & 36.4 & 38.1 & 36.8 & 38.9 & 37.9 & 30.7  \\
Gemini-2.5-Flash & 70.2 & 96.1 & 65.9 & 67.1 & 77.3 & 44.5 & 68.7 & 97.1 & 62.0 & 65.6 & 75.2 & 43.7 & 39.4 & 35.1 & 53.7 & 36.6 & 34.3 & 37.2 \\
Qwen3.6-plus & 68.3 & 95.2 & 57.0 & 70.3 & 80.9 & 38.2 & 68.9 & 96.7 & 58.9 & 70.6 & 77.4 & 40.8 & 34.0 & 33.7 & 39.8 & 33.2 & 28.0 & 35.1 \\
Doubao-Seed-2.0-lite & 69.5 & 98.9 & 64.0 & 69.3 & 79.2 & 36.3 & 66.6 & 98.8 & 62.1 & 66.3 & 75.1 & 30.8 & 42.4 & 52.8 & 50.4 & 36.6 & 41.4 & 31.2 \\
Kimi-K2.5  & 77.8 & 94.4 & \textbf{81.8} & 77.9 & 90.8 & 44.3 & 67.2 & 91.9 & 61.3 & 69.6 & 83.3 & 30.2 & 63.1 & 56.5 & \textbf{85.4} & 57.3 & 68.9 & 47.7  \\
GPT-5.5 & 76.0 & 99.8 & 65.4 & 71.1 & 91.7 & 52.0 & 74.7 & \textbf{99.9} & 62.3 & 67.7 & \textbf{92.2} & 51.2 & 44.5 & 42.8 & 57.6 & 42.8 & 38.6 & 40.9\\
Gemini-3.5-Flash & \textbf{83.9} & \textbf{99.9} & 68.3 & \textbf{87.3} & \textbf{94.9} & \textbf{69.1} & \textbf{78.7} & 99.7 & \textbf{62.7} & \textbf{78.5} & 90.8 & \textbf{61.8} & \textbf{70.3} & \textbf{63.6} & 82.5 & \textbf{69.2} & \textbf{73.7} & \textbf{62.6}  \\
Claude Opus 4.8 & 73.6 & 97.2 & 69.8 & 65.6 & 79.1 & 56.5 & 73.8 & 98.6 & 59.5 & 70.7 & 84.0 & 56.2 & 47.1 & 41.3 & 60.2 & 41.7 & 44.5 & 48.1 \\
\midrule
Human &  &  &  &  &  &  &  &  &  &  &  &  & 92.9 & 83.7 & 96.2 & 94.2 & 97.5 & 93.2 \\ 

\bottomrule
\end{tabular}
} 
\end{table*}

\textbf{(1) Visual Maps Serve as Critical External Representations for Enhancing Spatial Reasoning.} The results clearly show that \textbf{Data + Map} provides a significant advantage in accuracy, emphasizing that visual maps are essential tools for enhancing spatial reasoning, not merely decorative elements. On average, incorporating discrete choropleth maps improves accuracy by 6.0\% over \textbf{Data Only} and 14.8\% over \textbf{Map Only}. For continuous maps, the improvements are 3.9\% and 16.3\%, respectively. This improvement is especially pronounced in models with higher baseline accuracy, with models achieving over 60\% accuracy seeing an average boost of around 8\%. For instance, Kimi-K2.5 and Gemini-3.5-Flash, which achieve 77.8\% and 83.9 \% in the \textbf{Data + Map} condition, improve by 10.6\% and 5.2\% over their performance in the \textbf{Data Only} condition (67.2\% and 78.7\%). Similarly, InternVL3.5-8B sees a rise from 60.8\% to 64.9\% with the addition of maps. This boost highlights two important points: first, maps provide structured spatial representations that anchor numerical data in a geographical context, facilitating better spatial reasoning. Second, models with higher accuracy seem to benefit more from maps, as they already have a stronger data processing foundation, allowing maps to enhance their ability to integrate complex spatial relationships.

Conversely, the \textbf{Map Only} condition performed the worst, revealing a key limitation of current multimodal models: they struggle to decode quantified information from visual representations alone. While humans easily interpret numerical values from maps (with most task dimensions exceeding 90\%), models struggle with this visual quantification. This gap underscores the challenge in complex spatial reasoning, where models act more as symbol processors than visual experts. The core value of maps lies in their ability to provide spatial anchors that integrate symbolic data with geographical context, as seen from the improved performance when both visual and symbolic inputs are combined.

In summary, the results highlight that \textbf{Data + Map} significantly enhances model performance by integrating visual and symbolic information. Maps are crucial not just for visualization, but for supporting more effective spatial reasoning, particularly when paired with symbolic data. The cognitive advantage of maps can only be fully realized when visual perception is integrated with symbolic reasoning, as evidenced by the improvement in task performance across the board.

\textbf{(2) Performance Exhibits a Sharp Decay as Tasks Shift from Local Perception to Global Pattern Recognition.} A more granular analysis of the tasks reveals that models excel at local perception but struggle as tasks require more global reasoning. In the \textit{D1. Identify} tasks, models such as GPT-5.5 and Gemini-3.5-Flash achieve near-perfect accuracy scores of 99.8\% and 99.9\%, respectively. These tasks, which involve simple identification or classification of spatial features, are well within the capabilities of current models. However, when tasks progress to \textit{D5. Delineate}, which involves synthesizing information across the entire map and recognizing complex spatial patterns, performance drops sharply. Even the best-performing model, Gemini-3.5-Flash, experiences a significant drop to 69.1\% accuracy in D5. This sharp decline highlights a critical limitation of current models: while they excel at 'point-lookup' tasks (e.g., identifying specific locations or objects), they still lack the ability to maintain a `global vision' necessary for identifying spatial trends or anomalies across fragmented regions of a map. The performance collapse from D1 to D5 suggests that models are still developing the ability to integrate global context and infer relationships over larger geographic areas. In contrast, human evaluators achieve an impressive 93.2\% accuracy on D5, highlighting the human ability to synthesize complex spatial information across the entire map. This sharp performance gap between models and humans emphasizes the ongoing challenge in artificial intelligence for complex spatial reasoning.

\textbf{(3) Significant Gap Persists Between Machine Perception and Human Cartographic Interpretation.}
Despite the substantial advancements in proprietary models, there remains a significant gap between machine and human performance in the \textbf{Map Only} condition. Human evaluators achieved a remarkable 92.9\% accuracy, purely relying on their ability to interpret the visual cues of the map, such as color intensity and spatial proximity. In contrast, the best-performing model in this condition, Gemini-3,5-Flash, only achieved 70.3\%. Many open-source models, like those in the Qwen3-VL series, struggled even more, with accuracy hovering around 35\%, barely outperforming random chance.

Across different spatial reasoning tasks, model performance shows a non-monotonic pattern. The highest performance is observed in \textit{D2. Spatial Recognition}, where models reach 82.5\%, representing their closest alignment with human performance (96.2\%), while the lowest performance occurs in \textit{D5. Delineate}, where models drop to 62.6\% compared to 93.2\% for humans. In other tasks, models show moderate but consistently lower performance than humans, including D1 (63.6\% vs. 83.7\%), D3 (69.2\% vs. 94.2\%), and D4 (73.7\% vs. 97.5\%), indicating that despite partial improvements in intermediate reasoning, a substantial gap persists across all task types.

Overall, these results indicate that machine performance peaks at intermediate spatial reasoning (D2), but remains consistently below human performance across all tasks, with particularly large gaps in both low-level perception and high-level spatial integration.


\textbf{(4) Model Families and Limitation of Scaling.} The results also suggest that increasing model size significantly improves spatial reasoning capabilities. For instance, within the Qwen3.5 family, increasing the model's parameters from 2B to 9B results in a substantial improvement in performance. Qwen3.5-9B achieves an overall accuracy of 61.3\% in the \textbf{Data + Map} condition, a marked improvement from Qwen3.5-2B, which only reached 43.4\%. Similarly, in the InternVL3.5 series, the 8B version outperforms the 4B and 2B versions across almost all metrics, including overall accuracy and task-specific performance. This trend suggests that larger models are more capable of handling the complexities of cross-modal alignment, particularly in tasks that require synthesizing symbolic data with geographical maps. However, despite these improvements, current models still fall far short of human-level spatial reasoning. Even the largest models show a significant gap in performance compared to human evaluators, particularly in complex tasks requiring the integration of broad spatial contexts.

\subsection{Influence Analysis of Choropleth Map Design Factors}
Map design can vary significantly, and such diversity may influence the performance of foundation models. In our benchmark, we specifically considered various design factors, including map type, color hue, and spatial structure. Since the \textbf{Data + Map} setting consistently achieved the strongest overall performance in Table \ref{tab:overall-results}, all analyses in this section are conducted under this condition to reveal how internal map design factors affect model reasoning.

\subsubsection{Effect of Map Type: Discrete vs. Continuous Maps}


We compare 22 models' accuracy on discrete class-based maps and continuous value-based maps under the \textbf{Data + Map} condition. The aggregated results are shown in Figure~\ref{MapAnalysis}(a), indicating that discrete maps consistently outperform continuous maps across most evaluated models. Specifically, 21 out of 22 models performed better with discrete maps. Among the models, Qwen3.5 and Kimi-K2.5 show the largest differences. For instance, the Qwen3.5-9B model achieved 67.4\% accuracy with discrete maps, compared to just 55.3\% with continuous maps. Similarly, Kimi-K2.5 demonstrated a significant improvement, reaching 80.4\% with discrete maps versus 75.2\% with continuous maps. This can be attributed to the fact that discrete maps simplify the mapping between visual cues and semantic labels by converting continuous attribute distributions into categorical partitions. This discretization reduces ambiguity and functions as a form of information compression. In contrast, continuous maps retain the full numerical variability of attributes but introduce challenges in perceptual decoding, requiring models to interpret subtle color gradients. This often leads to degraded performance in tasks requiring exact comparisons or threshold-based reasoning.



Overall, these results highlight a fundamental trade-off between interpretability and informational fidelity in map types. Discrete maps enhance perceptual clarity and symbolic alignment, not only reducing human cognitive load but also easing the machine's decoding process. In contrast, continuous maps preserve richer quantitative structure, but at the cost of increased complexity in decoding and interpretation.

\subsubsection{Effect of Color Hue}


We compare the average accuracy of all evaluated models across 12 evenly spaced hue settings under the \textbf{Data + Map} condition. The aggregated results are shown in Figure~\ref{MapAnalysis}(b), indicating that model performance remains highly consistent across different hues, with only minor fluctuations observed. In most cases, the variation across hues is limited to only a few percentage points (\textless 2.5\%). This robustness likely arises because modern multimodal models rely more on relative spatial organization, class contrast, and region correspondence than on any specific semantic preference for individual colors. As long as sufficient visual contrast is preserved, models can generally interpret the map consistently regardless of hue choice.


Overall, these results suggest that foundation models are largely invariant to hue variation. Unlike human cartographic design, where hue may influence aesthetics, attention, or emotional associations, hue in this benchmark functions primarily as a low-level visual carrier with limited impact on machine spatial understanding.

\subsubsection{Effect of Spatial Structure}

\begin{figure}[H]
	\centering
	\includegraphics[width=\textwidth]{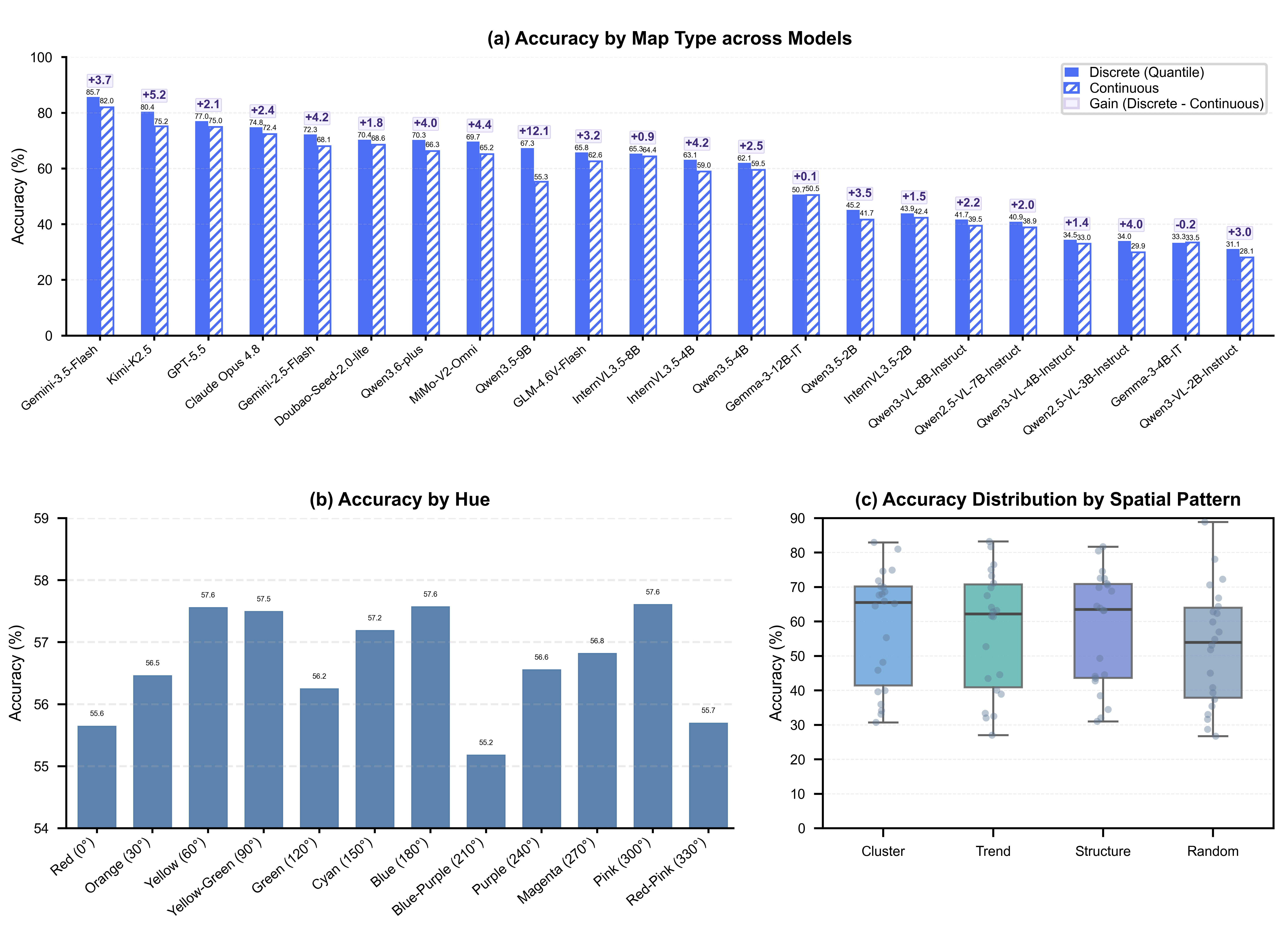}
	\caption{Different design factors' influence on models. (a) Accuracy by map type across models; (b) Accuracy by hue; (c) Accuracy distribution by spatial pattern.}
	\label{MapAnalysis}
\end{figure}

We compare the accuracy distributions of all evaluated models across four spatial structures under the \textbf{Data + Map} condition: Cluster, Trend, Structure, and Random. The aggregated results are shown in Figure~\ref{MapAnalysis}(c), indicating that organized spatial patterns generally yield higher performance than unstructured random patterns. Specifically, Cluster achieves the highest median accuracy (65.6\%), followed by Structure (63.5\%), while Trend remains competitive (62.2\%). In contrast, Random exhibits the lowest median performance (54.0\%), together with a generally lower upper range. This pattern suggests that foundation models benefit from coherent spatial regularities that can be visually summarized at a global level. Cluster maps contain contiguous high- or low-value regions, making it easier for models to detect grouping and neighborhood consistency. Trend maps provide directional gradients, which can be captured through monotonic changes across space. Structure maps, such as ring-like or nested forms, also provide meaningful organization, although they may require more complex global integration than simple clusters or trends. By comparison, Random maps lack salient large-scale regularities, forcing models to rely more heavily on local region-by-region inspection. This increases the reasoning burden and reduces the advantage of map-based visual abstraction.

Overall, these results suggest that maps are most helpful when the underlying data contain interpretable spatial structures. Foundation models can exploit organized global patterns, whereas purely random distributions offer fewer cues for efficient spatial reasoning.

\section{Discussion}
\subsection{Robustness to Prompting Strategies}

Recent studies have shown that prompting strategies can substantially influence the reasoning behavior and final performance of foundation models \cite{wei2025thinking}. Thus, we further examine the robustness of model performance under different prompting strategies. We compare three representative paradigms using Kimi-K2.5 (SOTA model as illustrated in Table \ref{tab:overall-results}): Direct Thought (DT), Chain-of-Thought (CoT), and Tree-of-Thought (ToT). All strategies are tested on the same benchmark tasks, while only the prompting format is varied, and all other input settings remain unchanged. Specifically:

\begin{itemize}
    \item \textbf{Direct Thought (DT):} a standard instruction-based prompt in which the model directly outputs the final answer without explicitly generating intermediate reasoning steps.
    
    \item \textbf{Chain-of-Thought (CoT):} the prompt explicitly instructs the model to reason step by step by adding the phrase ``Let's think step by step'' before producing the final answer.
    
    \item \textbf{Tree-of-Thought (ToT):} the prompt divides reasoning into three stages. First, the model generates a tentative reasoning plan for solving the spatial task. Second, conditioned on this plan, the model generates three candidate answers with step-by-step reasoning under a higher temperature (1.0) setting to encourage diversity. Third, all candidate responses are presented back to the model, which is asked to select the most plausible answer as the final prediction. To make the final selection more stable, this selection stage is conducted at a relatively low temperature of 0.1.
\end{itemize}

\begin{table}[htbp]
\centering
\small
\caption{Overall accuracy (\%) of Kimi-K2.5 using Direct Thought (DT), Chain-of-Thought (CoT), and Tree-of-Thought (ToT) prompting under the Data + Map, Data Only, and Map Only input conditions. All strategies are evaluated on the same benchmark questions with the remaining experimental settings held constant.}
\label{Prompt}
\begin{tabular}{l c c c}
\toprule
Prompt & Data+Map & Data Only & Map Only \\
\midrule
DT & 77.8 & 67.2 & 63.1  \\
CoT & 76.6 & 68.0 & 56.9  \\
ToT & 80.4 & 71.6 & 68.2  \\
\bottomrule
\end{tabular}
\end{table}

The experimental results are shown in Table~\ref{Prompt}, leading to the following observations: ToT consistently achieves the best performance across all input settings, outperforming DT by +2.57\%, +4.38\%, and +5.07\% in \textbf{Data+Map}, \textbf{Data Only}, and \textbf{Map Only} scenarios, respectively. This demonstrates that introducing multiple reasoning candidates through repeated sampling and structured selection significantly enhances model robustness. The improvement is particularly notable in the \textbf{Map Only} setting, suggesting that exploring multiple reasoning paths helps mitigate ambiguity in spatial structure interpretation.

CoT exhibits mixed and unstable performance. While it slightly improves the \textbf{Data Only} setting (+0.77\%), it leads to performance degradation in \textbf{Data+Map} (-1.21\%) and especially in \textbf{Map Only} (-6.22\%). This indicates that enforcing a simple or single linear reasoning chain may introduce unnecessary or even misleading intermediate steps, particularly when the task involves interpreting complex spatial distributions rather than purely numerical relationships. A possible explanation is that spatial reasoning from maps differs fundamentally from reasoning over textual or numerical information. In the \textbf{Data+Map} setting, the structured attribute values provide explicit numerical evidence that can continuously constrain and verify intermediate reasoning steps. Consequently, Chain-of-Thought (CoT) reasoning can effectively combine visual cues with numerical information to support more reliable inference. In contrast, under the \textbf{Map Only} setting, models must rely exclusively on visual color patterns to infer spatial relationships without access to the underlying attribute values. This process depends more heavily on holistic visual perception than on sequential symbolic reasoning. As a result, forcing the model to generate a linear reasoning chain may encourage it to over-interpret ambiguous visual evidence, introduce unsupported intermediate assumptions, and propagate early perception errors throughout the reasoning process. Rather than improving performance, the additional reasoning steps can therefore amplify incorrect visual interpretations, leading to the observed degradation in the \textbf{Map Only} condition.

Overall, these results suggest that prompting strategies provide useful but limited gains compared with input representation itself. For practical map reasoning tasks, improving the input representation (e.g., maps with data) is likely more beneficial than prompt engineering alone. When prompting is needed, ToT appears more reliable than simple step-by-step prompting. However, this gain comes with a higher cost, requiring about five times more tokens than DT. In practice, its modest accuracy improvement should be weighed against the additional computational and monetary expense. 

\subsection{Perceptual Biases from Prior Knowledge}
\subsubsection{Influence of Language.}
To examine whether input language affects model performance in geospatial reasoning tasks, we conduct a controlled experiment using Chinese and English prompts. We focus on these two languages because they represent the primary languages of the two most active regions in current FM development, namely the United States and China. Correspondingly, we select two representative frontier models: Gemini-2.5-Flash and Kimi-K2.5. Both models are evaluated under identical experimental settings, with only the prompt language varied. The results are shown in Table \ref{Language}.

\begin{table}[htbp]
\centering
\small
\caption{Overall accuracy (\%) of Gemini-2.5-Flash and Kimi-K2.5 using Chinese and English prompts under the Data + Map, Data Only, and Map Only input conditions. Only the prompt language is varied, while all other experimental settings remain unchanged.}
\label{Language}
\begin{tabular}{l l c c c}
\toprule
Model & Language & Data+Map & Data Only & Map Only \\
\midrule
\multirow{2}{*}{Gemini-2.5-Flash} & Chinese & 70.2 & 68.7 & 39.4  \\
                                   & English & 72.7 & 69.8 & 40.0  \\
\midrule
\multirow{2}{*}{Kimi-K2.5}        & Chinese & 77.8 & 67.2 & 63.1  \\
                                   & English & 79.7 & 69.1 & 66.9  \\
\bottomrule
\end{tabular}
\end{table}

The results in the table show that both models consistently achieve higher accuracy under English prompts across all input conditions. For Gemini-2.5-Flash, performance increases from 70.2\% to 72.7\% in \textbf{Data+Map}, from 68.7\% to 69.8\% in \textbf{Data Only}, and from 39.4\% to 40.0\% in \textbf{Map Only}. For Kimi-K2.5, the gains are more pronounced, improving from 77.8\% to 79.7\% in \textbf{Data+Map}, from 67.2\% to 69.1\% in \textbf{Data Only}, and from 63.1\% to 66.9\% in \textbf{Map Only}. Across input conditions, the language effect is consistently smaller in the Data+Map setting compared to Data Only and Map Only. This suggests that the presence of visual map information can partially reduce the reliance on linguistic understanding, thereby mitigating language-induced performance differences. Overall, these findings indicate that although modern multimodal models demonstrate a degree of cross-lingual robustness, performance still benefits from English prompts, likely reflecting the dominance of English in training data. Furthermore, the integration of visual information plays a stabilizing role by reducing language sensitivity.

\subsubsection{Influence of Prior knowledge}

The benchmark maps used in previous sections are generated from county-level administrative geometries of multiple countries, with region names replaced by neutral synthetic labels (e.g., two-letter codes), thereby minimizing semantic priors and emphasizing reasoning over map content. However, real-world models may still possess substantial prior knowledge about familiar places, especially widely represented countries such as the United States. Therefore, we further examine whether such implicit prior knowledge influences model performance by conducting an additional experiment on U.S.-based maps. Specifically, we regenerate the benchmark using the same pipeline described in \textbf{Section 3.2}, but replace the synthetic anonymized maps with real U.S. administrative units and realistic region names. The evaluation results under InterVL3.5B and Kimi-K2.5 are shown in Table \ref{Basemap}.

\begin{table}[htbp]
\centering
\small
\captionsetup{justification=centering,singlelinecheck=false}  
\caption{Overall accuracy (\%) of InternVL3.5-8B and Kimi-K2.5 on synthetic benchmark maps and maps based on real U.S. administrative units under the Data + Map, Data Only, and Map Only input conditions. The comparison evaluates the sensitivity of model performance to geographic context and realistic region names. }
\label{Basemap}
\begin{tabular}{l l c c c}
\toprule
Model & Basemap & Data+Map & Data Only & Map Only \\
\midrule
\multirow{2}{*}{InterVL3.5-8B} & Synthetic & 64.9 & 60.8 & 46.9  \\
                                & American  & 60.4 & 60.4 & 52.7  \\
\midrule
\multirow{2}{*}{Kimi-K2.5}     & Synthetic & 77.8 & 67.2 & 63.1  \\
                                & American  & 84.1 & 76.2 & 71.9  \\
\bottomrule
\end{tabular}
\end{table}

The results reveal notable model differences. Performance on InternVL3.5-8B decreases in the \textbf{Data+Map} condition from 64.9\% to 60.4\%, and slightly decreases in \textbf{Data Only} from 60.8\% to 60.4\%, while improving in \textbf{Map Only} from 46.9\% to 52.7\%. This suggests that the open-source model gains limited benefit from real-world geographic priors and may even be distracted by more complex authentic place semantics. By contrast, Kimi-K2.5 shows substantial gains across all settings. Accuracy rises from 77.8\% to 84.1\% in \textbf{Data+Map}, from 67.2\% to 76.2\% in \textbf{Data Only}, and from 63.1\% to 71.9\% in \textbf{Map Only}. These improvements of +6.3, +9.0, and +8.8 percentage points, respectively, indicate that proprietary frontier models may possess stronger prior geographic knowledge about well-known U.S. regions, administrative structures, or commonly seen spatial patterns.

Overall, these findings suggest that prior knowledge can significantly affect benchmark performance, but the magnitude depends strongly on model family and pretraining exposure. For controlled evaluation of pure spatial reasoning, anonymized synthetic maps remain important because they reduce memorization effects. At the same time, real-world maps provide a complementary setting for testing how practical geographic knowledge interacts with map understanding.
\subsection{Sensitivity Analysis}
\subsubsection{Temperature Settings}

Decoding temperature controls the randomness of token generation and may influence the reasoning process of foundation models. Although all main experiments in this study adopt greedy decoding ($T=0.0$), it remains unclear whether the reported findings are sensitive to different decoding temperatures. Therefore, we further evaluate two representative models, InternVL3.5-8B and Gemini-3.5-Flash, under a range of temperature settings while keeping all other experimental configurations unchanged.

The results are summarized in Table~\ref{temp}. Overall, both models exhibit relatively stable performance under different temperature settings, indicating that the conclusions of this study are not sensitive to moderate decoding randomness.

\begin{table}[htbp]
\centering
\small
\captionsetup{justification=centering,singlelinecheck=false}  
\caption{Overall accuracy (\%) of InternVL3.5-8B and Gemini-3.5-Flash at different decoding temperatures under the Data + Map, Data Only, and Map Only input conditions. Only the decoding temperature is varied, while all other experimental settings remain unchanged. A temperature of 0.0 denotes greedy decoding. }
\label{temp}
\begin{tabular}{l c c c c}
\toprule
Model & Temperature & Data+Map & Data Only & Map Only \\
\midrule
\multirow{6}{*}{InterVL3.5-8B} & 0.0 & 64.9 & 60.8 & 46.9  \\
                                & 0.2 & 64.7 & 61.1 & 46.8 \\
                                & 0.4 & 64.7 & 60.6 & 46.7 \\
                                & 0.6 & 64.0 & 60.4 & 45.9 \\
                                & 0.8 & 63.0 & 60.1 & 45.5\\
                                & 1.0 & 62.5 & 58.7 & 45.2\\
\midrule
\multirow{6}{*}{Gemini-3.5-Flash}   & 0.0 & 83.9 & 78.7 & 70.3  \\
                                & 0.2 & 83.3 & 78.3 & 67.9 \\
                                & 0.4 & 83.3 & 78.0 & 67.6\\
                                & 0.6 & 83.0 & 78.0 & 67.5\\
                                & 0.8 & 83.1 & 77.8 & 67.0\\
                                & 1.0 & 82.7 & 77.1 & 66.5\\
\bottomrule
\end{tabular}
\end{table}

For InternVL3.5-8B, greedy decoding ($T=0.0$) achieves the highest accuracy under all three input settings. As the temperature increases, performance gradually declines. Specifically, accuracy in the \textbf{Data+Map} setting decreases from 64.9\% to 62.5\%, while the \textbf{Data Only} and \textbf{Map Only} settings decrease from 60.8\% to 58.7\% and from 46.9\% to 45.2\%, respectively. Although the degradation is consistent, the overall change remains modest (within 2.4 percentage points), suggesting that the model's spatial reasoning capability is relatively insensitive to decoding temperature.

A similar trend is observed for Gemini-3.5-Flash. Across all three input settings, the best performance is again obtained with greedy decoding, and increasing the temperature results in only slight reductions in accuracy. Even at $T=1.0$, the performance decreases by only 1.2, 1.6, and 3.8 percentage points under the \textbf{Data+Map}, \textbf{Data Only}, and \textbf{Map Only} settings, respectively. Compared with InternVL3.5-8B, Gemini-3.5-Flash demonstrates slightly greater robustness to decoding randomness, particularly in the \textbf{Data+Map} and \textbf{Data Only} settings.

These observations indicate that decoding temperature has only a limited influence on benchmark performance. More importantly, the relative performance trends across different input settings remain highly consistent under all temperature configurations. In particular, the superiority of the \textbf{Data+Map} setting over \textbf{Data Only} and \textbf{Map Only} is preserved regardless of the decoding temperature, and the ranking between representative models also remains unchanged. This demonstrates that the conclusions drawn in this paper are robust to decoding randomness rather than being artifacts of a particular temperature setting. Therefore, all main experiments adopt greedy decoding ($T=0.0$) to ensure reproducibility while preserving the same empirical conclusions.

\subsubsection{Quantization Levels}

Low-bit quantization is widely adopted to reduce GPU memory consumption and accelerate inference for large foundation models. In this work, all open-source models are evaluated using 4-bit NF4 quantization, which provides an efficient balance between memory usage and inference accuracy. To examine whether the conclusions of this study depend on the adopted quantization configuration, we further compare three representative precision settings, namely BF16 (without quantization), 4-bit NF4 quantization, and 8-bit integer quantization (INT8), while keeping all other experimental settings unchanged.

\begin{table}[htbp]
\centering
\small
\captionsetup{justification=centering,singlelinecheck=false}  
\caption{Overall accuracy (\%) of InternVL3.5-8B with NF4 4-bit quantization, FP16 precision, and BF16 precision under the Data + Map, Data Only, and Map Only input conditions. All other experimental settings remain unchanged.}
\label{quant}
\begin{tabular}{c c c c}
\toprule
Quantization & Data+Map & Data Only & Map Only \\
\midrule
NF4 & 64.9 & 60.8 & 46.9 \\
INT8 & 65.6 & 61.4 & 48.7 \\
BF16 & 66.3 & 61.5 & 49.1 \\

\bottomrule
\end{tabular}
\end{table}

The results are summarized in Table~\ref{quant}. Overall, increasing numerical precision leads to only modest improvements in benchmark performance. For InternVL3.5-8B, the overall accuracy under the \textbf{Data+Map} setting increases from 64.9\% with 4-bit NF4 quantization to 65.6\% with INT8 and 66.3\% with BF16. Similar but slightly smaller improvements are observed under the \textbf{Data Only} and \textbf{Map Only} settings, where BF16 achieves the highest accuracy, followed by INT8 and NF4.

These results indicate that reducing quantization error can slightly improve spatial reasoning performance, particularly in the visually challenging \textbf{Map Only} setting, where the performance gain reaches 2.2 percentage points from NF4 to BF16. Nevertheless, the overall differences remain relatively small, suggesting that the benchmark results are not highly sensitive to the choice of quantization precision.

More importantly, the relative performance trends remain unchanged across different precision settings. In particular, the \textbf{Data+Map} setting consistently outperforms \textbf{Data Only}, \textbf{Map Only}, and the overall conclusions of this paper are preserved regardless of whether the model is evaluated using NF4 quantization, INT8 quantization, or full-precision BF16 inference. Considering its substantially lower GPU memory consumption and competitive performance, NF4 quantization provides an effective trade-off between computational efficiency and evaluation accuracy, and is therefore adopted throughout the main experiments.

\subsubsection{Data Distribution and Classification Methods}

The benchmark attributes are sampled from real-world statistical variables rather than synthetically generated values. To better characterize the benchmark, we first analyze the distributions of all attribute values. The results show that skewed distributions account for 87.4\% of all benchmark instances, whereas approximately normal, multimodal, and nearly uniform distributions account for 6.8\%, 3.3\%, and 2.5\%, respectively. This observation is consistent with many real-world socioeconomic indicators, which commonly exhibit non-uniform and long-tailed distributions.

Although the benchmark reflects realistic attribute distributions, choropleth map interpretation depends not only on the underlying data but also on how the data are classified into color classes\cite{Brewer1994}. Different classification methods may produce substantially different visual patterns even when the underlying values remain unchanged, potentially affecting the spatial reasoning of foundation models. To investigate this effect, we compare three widely used choropleth classification methods, namely Equal Interval, Quantile, and Natural Breaks \cite{yang2025mapcolorai}. We restrict the analysis to D1, D3, and D4 because these tasks directly evaluate the mapping between attribute values and color encoding while keeping the underlying reasoning objective unchanged. In contrast, changing the classification method may fundamentally alter the visual spatial patterns represented in D5, requiring regeneration of both the benchmark questions and the corresponding ground-truth answers. D2 focuses primarily on spatial relationships, which are not directly affected by the classification strategy. Therefore, D2 and D5 are excluded from this controlled comparison. The geographic regions and underlying attribute values are kept identical, while only the classification method is changed, allowing the influence of choropleth symbolization on foundation model spatial reasoning to be evaluated independently.

\begin{table}[htbp]
\centering
\small
\caption{Overall accuracy (\%) of InternVL3.5-8B using the original benchmark classification, Equal Interval, Quantile, and Natural Breaks classification methods under the Data + Map, Data Only, and Map Only input conditions. The analysis includes D1 (Identify), D3 (Compare), and D4 (Rank), whose answers can be recalculated independently of the spatial-pattern-generation procedure.}
\label{class}
\begin{tabular}{l c c c}
\toprule
Classification Methods & Data+Map & Data Only & Map Only \\
\midrule
Original & 77.7 & 75.0 & 49.4  \\
Equal Interval & 76.7 & 74.0 & 51.9  \\
Quantile & 77.4 & 73.4 & 52.1\\
Natural Breaks & 68.4 & 62.3 & 51.6 \\
\bottomrule
\end{tabular}
\end{table}

The results are summarized in Table~\ref{class}. Overall, Equal Interval and Quantile produce performance that is highly consistent with the original benchmark. For InternVL3.5-8B, the \textbf{Data+Map} accuracy changes by less than 1.0 percentage point under both classification methods (77.7\% vs. 76.7\% and 77.4\%), while the \textbf{Data Only} and \textbf{Map Only} settings exhibit similarly small variations. These observations suggest that moderate changes in choropleth classification have only a limited influence on the model's spatial reasoning ability.

In contrast, Natural Breaks leads to a noticeable performance degradation in the \textbf{Data+Map} and \textbf{Data Only} settings, where the accuracy decreases to 68.4\% and 62.3\%, respectively. This reduction is expected because Natural Breaks optimizes class boundaries according to the data distribution, producing substantially different color intervals from the original benchmark. Consequently, some ordinal relationships that are visually preserved under Equal Interval or Quantile become compressed or expanded, making value interpretation more difficult for the model. Interestingly, the \textbf{Map Only} accuracy remains relatively stable across all three classification methods (49.4\%--52.1\%), suggesting that when numerical values are unavailable, changing the classification strategy alone has only a limited effect on purely visual spatial reasoning.

Overall, these results indicate that choropleth classification methods do influence the spatial reasoning performance of FMs. Specifically, the \textbf{Data+Map} setting consistently achieves the highest performance, followed by \textbf{Data Only} and \textbf{Map Only}, regardless of how the choropleth map is classified. Therefore, while the classification strategy affects the absolute performance of FMs, it does not alter the relative effectiveness of different input modalities or the main conclusions of this study.

\subsection{Stability and Repeatability of Model Responses}

To evaluate the stability and repeatability of model performance, we conduct repeated inference experiments under identical experimental settings. Specifically, we select one representative open-source model (InternVL3.5-8B) and one proprietary model (Kimi-K2.5), and run each model three times on the full benchmark under 3 conditions. All runs adopt the same prompt template and decoding configuration, ensuring that any observed variation is attributable to intrinsic model stochasticity rather than external factors. Table~\ref{Repeatability} reports the overall accuracy performance across the three runs. 

\begin{table}[htbp]
\centering
\small
\caption{Overall accuracy (\%) of InternVL3.5-8B and Kimi-K2.5 across three repeated runs under the Data + Map, Data Only, and Map Only input conditions. All runs use identical prompts, decoding configurations, and benchmark data, allowing response stability and repeatability to be assessed. }
\label{Repeatability}
\begin{tabular}{l l c c c}
\toprule
Model & Run & Data+Map & Data Only & Map Only \\
\midrule
\multirow{3}{*}{InternVL3.5-8B} & Run 1 & 64.9 & 60.8 & 46.9  \\
                                 & Run 2 & 64.9 & 60.8 & 46.9  \\
                                 & Run 3 & 64.9 & 60.8 & 46.9  \\
\midrule
\multirow{3}{*}{Kimi-K2.5}      & Run 1 & 77.8 & 67.2 & 63.1  \\
                                 & Run 2 & 77.8 & 67.3 & 62.6  \\
                                 & Run 3 & 77.3 & 67.3 & 62.7  \\
\bottomrule
\end{tabular}
\end{table}





For InternVL3.5-8B, the results are fully consistent across all three runs, with identical accuracy values under every input condition. This indicates that the model behaves deterministically under the current evaluation setup and that its benchmark results are highly stable. For Kimi-K2.5, slight variations are observed across repeated runs, but the fluctuations remain minimal. Accuracy varies only from 77.3\% to 77.8\% in \textbf{Data+Map}, from 67.2\% to 67.3\% in \textbf{Data Only}, and from 62.6\% to 63.1\% in \textbf{Map Only}. The maximum deviation across runs is no more than 0.5 percentage points, indicating strong consistency despite minor stochasticity in API-based inference. Importantly, these run-to-run variations are negligible compared with the performance differences between input conditions. For example, Kimi-K2.5 shows a gap of more than 10 percentage points between \textbf{Data+Map} (77.8\%) and \textbf{Data Only} (67.2\%), which is far larger than the repeated-run fluctuation.

Overall, both models demonstrate high stability and repeatability. This suggests that the experimental results reported in this study are reliable and that the main conclusions are not driven by random inference noise.

\subsection{Why Do Maps Still Matter for Machines: Case Analysis}

\begin{figure}[H]
	\centering
	\includegraphics[width=\textwidth]{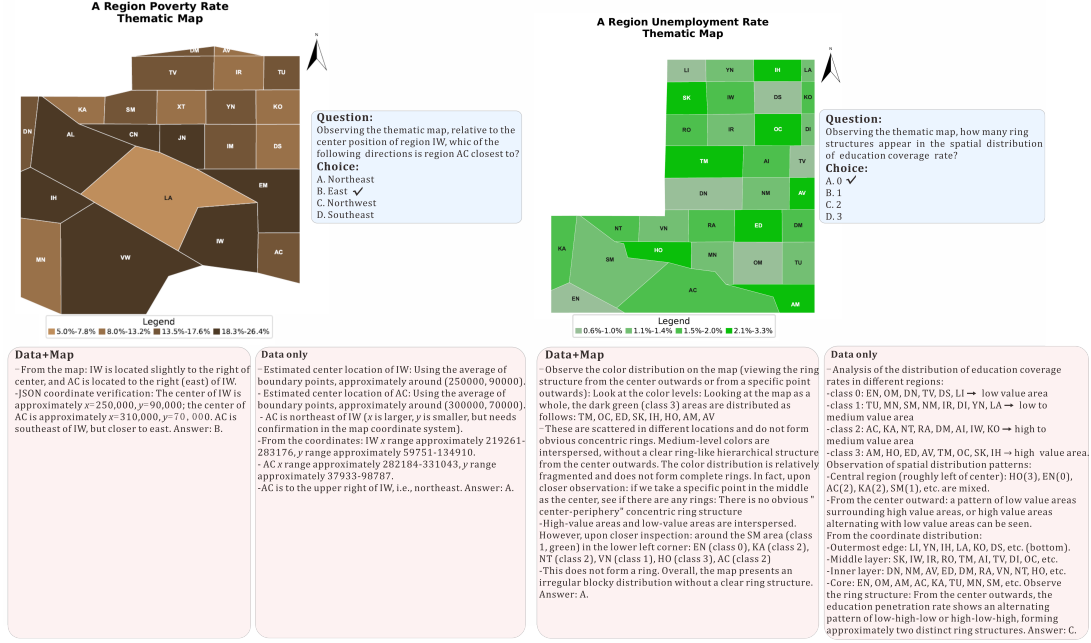}
	\caption{Two representative cases illustrating how map visualization supports spatial reasoning under the Data + Map condition compared with the Data Only condition. The left case concerns local directional-relation reasoning, whereas the right case concerns recognition of a global spatial trend. For each case, the choropleth map, benchmark question, model responses under the two input conditions, and corresponding reasoning processes are presented.}
	\label{CaseAnalysis}
\end{figure}

To further understand why maps remain beneficial for machine reasoning, we analyze two representative cases from our benchmark (Case 1: D2; Case 2: D5), where models succeed under the \textbf{Data + Map} condition but fail under \textbf{Data Only}, as shown in Figure \ref{CaseAnalysis}.

\textbf{Case 1: Local spatial relation reasoning (D2).}  
D2 requires identifying the relative direction of one region with respect to another. Under \textbf{Data + Map}, the model directly uses the visual layout to infer east--west and north--south relations, and then uses coordinates only as secondary verification. The reasoning is short, stable, and correct. In contrast, under \textbf{Data Only}, the model must reconstruct positions from boundary coordinates and estimated centroids, leading to a longer and more error-prone process that ultimately produces the wrong direction.

\textbf{Case 2: Global spatial pattern recognition (D5).}  
D5 asks the model to determine how many ring-like structures exist in the thematic map. Under \textbf{Data + Map}, the model can directly inspect the overall color arrangement and correctly recognize that no clear ring structure is present. Under \textbf{Data Only}, however, the model attempts to infer global structure from lists of values and approximate coordinates, which leads to over-interpreting noisy patterns as meaningful rings and produces an incorrect answer.

These two cases illustrate a consistent advantage of maps: they provide an explicit spatial reference frame that allows models to reason directly over arrangement, continuity, and global structure, rather than reconstructing space indirectly from raw coordinates and attributes. In this sense, maps function as cognitive compression interfaces, reducing reasoning burden, shortening reasoning chains, lowering token usage, and improving both accuracy and stability.

\subsection{Limitation and future work}

In this study, we provide empirical evidence that choropleth maps still offer clear benefits for FMs, especially when visual maps are combined with symbolic data for spatial understanding. Through a controlled benchmark comparing Data Only, Map Only, and Data + Map conditions across multiple frontier models, we show that maps remain effective external representations for machine intelligence. Nevertheless, several limitations should be acknowledged and point to important future directions. 

First, our benchmark focuses on choropleth maps as a controlled testbed. While choropleth maps are one of the most widely used thematic map types, many other cartographic representations, such as proportional symbol maps, flow maps, contour maps, cartograms, and interactive web maps, were not considered. Future work should extend the analysis to a broader range of map forms to examine whether the observed benefits of cartographic abstraction generalize beyond choropleth maps.

Second, although our benchmark systematically controls map type, hue, and spatial structure, the generated maps remain simplified synthetic environments. Real-world maps often contain richer geographic semantics, irregular layouts, noise, uncertainty, annotations, and mixed visual variables. Evaluating foundation models on authentic operational maps would provide stronger evidence of practical utility. In addition, while our study does not explicitly vary geographic regions (e.g., different countries), we acknowledge that pretrained foundation models may encode region-specific prior knowledge, which could influence model performance and generalization, and we leave systematic cross-region analysis as future work.

Third, our evaluation mainly uses accuracy on multiple-choice reasoning tasks. While this controlled setting enables fair comparison, it does not fully capture open-ended geographic analysis, decision support, interactive reasoning, or long-horizon planning. Future benchmarks could incorporate free-form responses, tool-augmented reasoning, and human-in-the-loop tasks to better reflect realistic applications.

Fourth, we evaluate current frontier multimodal models, but foundation models continue to evolve rapidly. Performance gaps observed today may narrow with stronger visual encoders, larger context windows, or specialized geospatial training. Continuous re-evaluation will therefore be necessary as new generations of models emerge.

Finally, our findings suggest that maps remain valuable external representations for machines, but the mechanisms behind this advantage are still not fully understood. Future work may explore how maps function as cognitive compression interfaces for AI, how map design can be optimized for machine readability, and whether new forms of AI-oriented cartography should be developed specifically for foundation models.

\section{Conclusion}

Recent advances in FMs have greatly strengthened machines’ ability to process raw multimodal inputs such as text, images, and structured data. In this context, an important question arises: if machines can increasingly reason directly from raw information, do traditional human-designed abstractions such as maps still matter? To address this question, we revisited the role of choropleth maps in FM spatial understanding through a controlled benchmark. We constructed ChoroplethMap-Bench, which systematically compares three input conditions: \textbf{Data Only}, \textbf{Map Only}, and \textbf{Data + Map}. The benchmark covers multiple map design factors, diverse spatial structures, and hierarchical reasoning tasks ranging from local attribute retrieval to global spatial pattern understanding. We evaluated a broad set of frontier open-source and proprietary multimodal models under a unified experimental protocol. The experimental results lead to several key findings. First, maps still provide clear benefits for machines. Across most evaluated models, the \textbf{Data + Map} condition consistently achieves the best performance, indicating that cartographic representations remain effective external interfaces for spatial reasoning. Second, the value of maps becomes more evident in higher-level tasks requiring global pattern recognition, where raw symbolic records alone impose a greater reasoning burden. Third, map design also matters for machines: discrete choropleth maps generally outperform continuous maps, while hue has limited influence under controlled settings. Fourth, maps additionally help reduce sensitivity to language variation and produce stable gains across repeated runs.

Overall, our findings suggest that maps are not obsolete in the era of FMs. Rather than being replaced by raw data processing, maps continue to function as cognitive compression tools that transform complex geographic information into more interpretable forms for both humans and machines. This work provides empirical evidence for the continued relevance of cartographic thinking in artificial intelligence and opens new opportunities for future research on AI-oriented map design, machine-readable cartography, and spatial intelligence through external representations.

\section*{Funding details}

This work was supported by grants from the National Natural Science Foundation of China (No. 42501551, 42371455) and the Tobii China Innovation Initiative Project (TPI250407CN).

\section*{Disclosure statement}

No potential conflict of interest was reported by the author(s).

\section*{Data availability statement}

The datasets and code used in this study are available at GitHub: https://github.com/Myantion/ChoroplethMap-Bench \cite{choroplethmap-bench2026}. The data and code are freely accessible and can be used for the purpose of reproducing the results (CC BY 4.0).

\noindent \textbf{Citation for the data and code}: Wei, Z., Sun, Y., Liu, Z., Xu, W., Dong, W., Liu, C., Liao, H., 2026.
ChoroplethMap-Bench: A benchmark for evaluating the cognitive
compression and spatial reasoning of MLLMs using 2,400 synthetic
choropleth maps. https://github.com/Myantion/ChoroplethMap-Bench. 

\section*{CRediT authorship contribution statement}
\textbf{Zhiwei Wei}: Conceptualization, Methodology, Investigation, Writing - Original draft preparation, Supervision, Project administration, Funding Acquisition. 
\textbf{Yonghe Sun}: Methodology, Data curation, Software, Validation, Visualization, Writing - Original draft preparation. 
\textbf{Zhenjia Liu}: Methodology, Data curation, Software, Visualization, Writing – review editing. 
\textbf{Wenjia Xu}: Writing - Review \& Editing, Supervision. 
\textbf{Chao He}: Conceptualization, Visualization. 
\textbf{Weihua Dong}: Writing - Review \& Editing, Supervision. 
\textbf{Chunbo Liu}: Writing - Review \& Editing, Supervision. 
\textbf{Hua Liao}: Writing - Review \& Editing, Supervision, Project administration, Funding Acquisition.

\clearpage
\appendix

\section{Prompts Used in the Experiments}
\label{appendix:prompts}

The complete prompts used in all experimental settings are provided below.

\subsection{Data + Map}
\begin{verbatim}
You are given a choropleth map together with its corresponding JSON data.
Please answer the following five multiple-choice questions by jointly
using the map and the JSON information.
For each question, output ONLY the correct option letter (A/B/C/D)
without any explanation.
JSON: {JSON}
Q1. {Question 1}
Options: {Options 1}
...
Q5. {Question 5}
Options: {Options 5}
The final output format must be exactly:
Q1: X
Q2: X
Q3: X
Q4: X
Q5: X
\end{verbatim}

\subsection{Data Only}
\begin{verbatim}
You are given a JSON data block.
Please answer the following five multiple-choice questions strictly
based on the JSON content.
For each question, output ONLY the correct option letter (A/B/C/D)
without any explanation.
JSON: {JSON}
Q1. {Question 1}
Options: {Options 1}
...
Q5. {Question 5}
Options: {Options 5}
The final output format must be exactly:
Q1: X
Q2: X
Q3: X
Q4: X
Q5: X
\end{verbatim}

\subsection{Map Only}
\begin{verbatim}
You will be shown a thematic map.
Answer the following five multiple-choice questions based only on the map.
For each question, output ONLY the correct option letter (A/B/C/D)
without any explanation.
Q1. {Question 1}
Options: {Options 1}
...
Q5. {Question 5}
Options: {Options 5}
The final output format must be exactly:
Q1: X
Q2: X
Q3: X
Q4: X
Q5: X
\end{verbatim}

\subsection{Direct Thought (DT)}
DT uses the corresponding prompt shown in \textbf{Sections A.1--A.3} without any additional reasoning instruction.

\subsection{Chain-of-Thought (CoT)}

The CoT prompt is identical to the corresponding baseline prompt except that the below instruction is inserted immediately before the model generates the final answer:

\begin{verbatim}
Let's think step by step.
\end{verbatim}

\subsection{Tree-of-Thought (ToT)}
The ToT prompting strategy consists of three sequential stages: planning, candidate generation, and selection.

\textbf{Stage 1 (Planning):} The following prompt is prepended to the corresponding baseline prompt to
generate a concise reasoning plan:

\begin{verbatim}
Please first provide a concise reasoning plan for answering the following
five questions.
Only output the reasoning plan. Do not provide the final answer choices.
\end{verbatim}

\textbf{Stage 2 (Candidate Generation):} The baseline prompt, together with the generated reasoning plan, is then
used to produce three independent candidate responses under
temperature = 1.0.

\textbf{Stage 3 (Selection):} The following prompt is prepended to the corresponding baseline prompt, followed by the three candidate responses:

\begin{verbatim}
You will be presented with multiple candidate answers to the same
questions. Please select the most reliable one and provide the final
answers.
Judge the consistency between the questions and the candidate responses,
and output the most reliable answers for Q1-Q5.
\end{verbatim}

The candidate responses are then appended after the following heading:

\begin{verbatim}
Candidate responses: {Candidate Responses}
\end{verbatim}

\subsection{Retry}

Based on the original prompt, an additional prompt is added for retry in case of invalid output:
\begin{verbatim}
[Strict output requirements] Your previous output format was not qualified.
\end{verbatim}

\end{document}